\documentclass[10pt,twocolumn,letterpaper]{article}

\usepackage{cvpr}              

\usepackage{makecell}

\definecolor{cvprblue}{rgb}{0.21,0.49,0.74}
\usepackage[pagebackref,breaklinks,colorlinks,allcolors=cvprblue]{hyperref}

\def\paperID{28} 
\def\confName{CVPR}
\def\confYear{2026}

\title{Improving Visual Grounding in Remote Sensing via Cluster-Guided Refinement and Model Ensemble Voting}

\author{Panav Shah\\
Indian Institute of Technology Bombay\\
Mumbai, India\\
{\tt\small panav.shah@iitb.ac.in}
\and
Geet Sethi\\
Indian Institute of Technology Bombay\\
Mumbai, India\\
{\tt\small 23b2258@iitb.ac.in}
\and
Ashutosh Gandhe\\
Indian Institute of Technology Bombay\\
Mumbai, India\\
{\tt\small 22b2409@iitb.ac.in}
}

\begin{document}
\maketitle
\begin{abstract}
Visual grounding aims to locate image regions that correspond to natural language descriptions and is a key component of interpretable vision systems. In remote sensing imagery, grounding is particularly challenging due to complex scenes, small objects, and large variations in scale. Relying on a single model is often insufficient to address these diverse challenges. In this work, we propose two grounding pipelines, Sequential Grounding Refinement (SGR) and Cluster-Aware Grounding Refinement (CGR), that combine the complementary strengths of RemoteSAM, a visual grounding model specialized for remote sensing, and SAM3, a powerful general-purpose segmentation model. Our approach first uses RemoteSAM to obtain an initial estimate of object location, which is then refined using SAM3 to produce more accurate and spatially consistent segmentations. Additionally, we explore an ensemble strategy based on majority voting across six diverse grounding pipelines, each with distinct capabilities. This multi-model framework improves robustness and significantly enhances localization accuracy. Experimental results demonstrate that the proposed pipelines and ensemble approach outperform individual models, leading to more reliable and precise visual grounding predictions.
\end{abstract}    
\section{Introduction}

Remote sensing imagery plays a vital role in a wide range of applications, including urban planning, disaster response, environmental monitoring, and infrastructure analysis. With the increasing availability of high-resolution satellite data, there is a growing need for automated methods that allow users to efficiently analyze and interpret large-scale geospatial imagery. Visual grounding, the task of localizing objects in an image based on natural language descriptions, provides a powerful solution by enabling intuitive, language-driven querying of remote sensing images.

\begin{figure}[t]
    \centering
    \begin{subfigure}{0.48\linewidth}
        \centering
        \includegraphics[width=\linewidth]{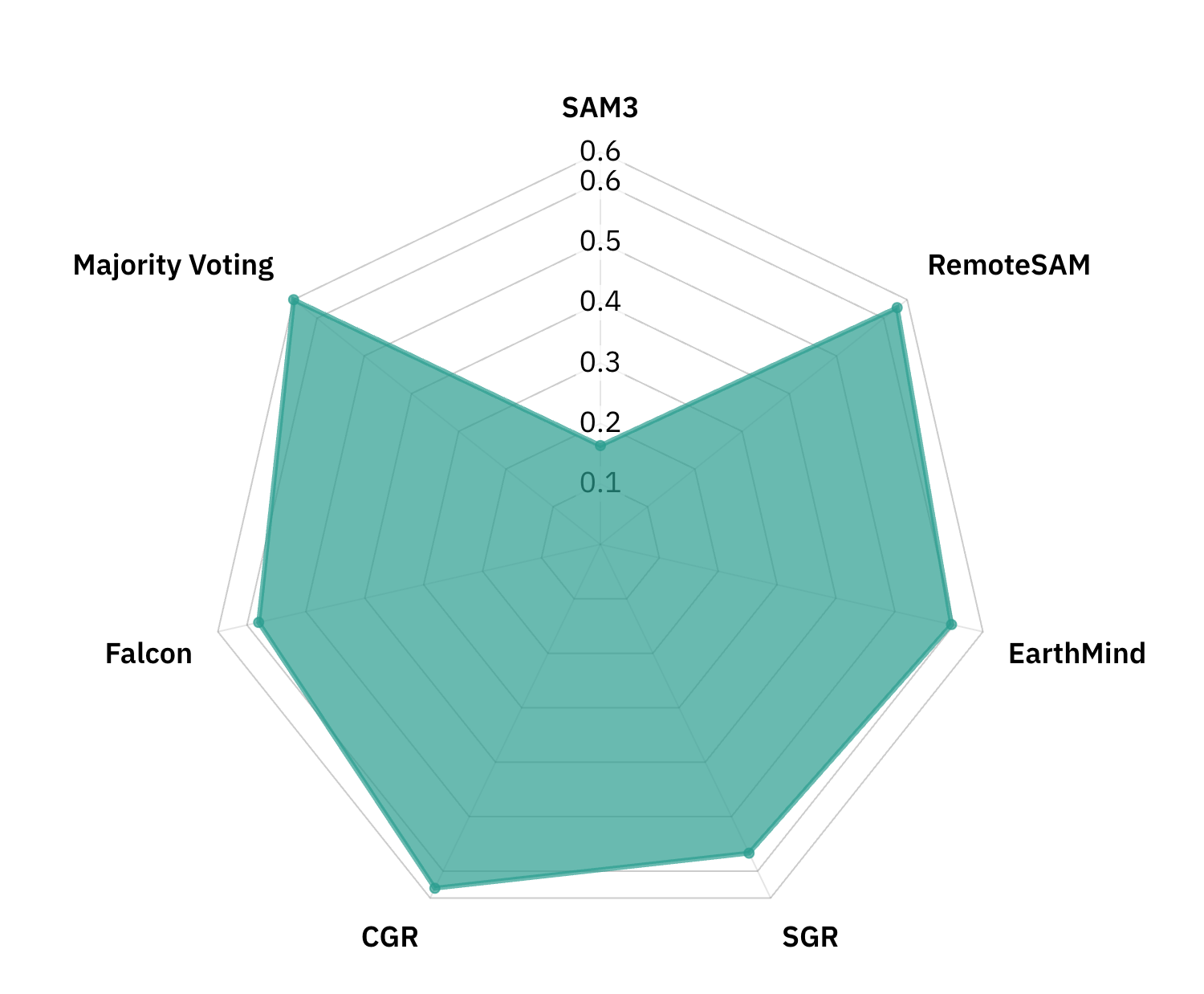}
        \caption{VRS Bench}
        \label{remotesam_sam3_v1}
    \end{subfigure}
    \hfill
    \begin{subfigure}{0.48\linewidth}
        \centering
        \includegraphics[width=\linewidth]{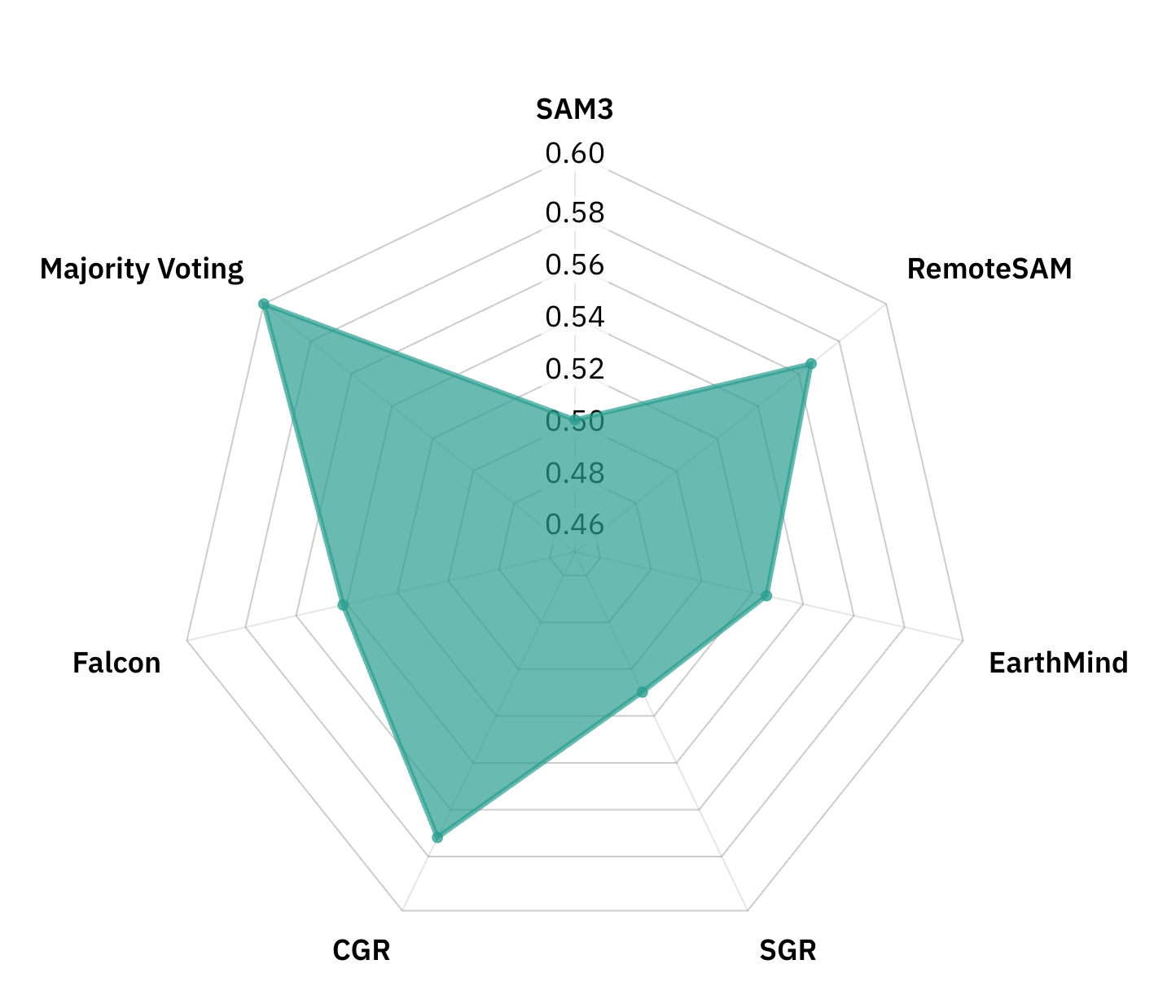}
        \caption{NWPU-VHR-10}
        \label{remotesam_sam3_v2}
    \end{subfigure}

    \caption{mIoU performance comparison of different models on the VRS Bench and NWPU-VHR-10 datasets. The proposed Cluster-Aware Grounding Refinement (CGR) pipeline and the majority-voting ensemble consistently outperform the individual models.}
    \label{diagrams}
\end{figure}

\begin{figure*}[t]
    \centering
    \includegraphics[width=1\linewidth]{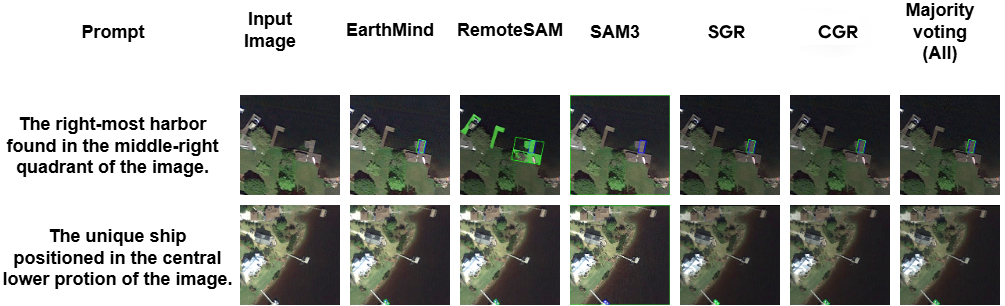}
    \caption{Qualitative comparison of grounding results from different methods.}
    \label{fig:grounding_pipelines}
\end{figure*}

However, grounding in remote sensing imagery is particularly challenging due to cluttered backgrounds, wide variations in object scale, and the presence of multiple visually similar targets within a single scene. Although several strong models for visual grounding and segmentation have recently been developed, each has inherent limitations. RemoteSAM \citep{yao2025remotesam}, a grounding model specialized for remote sensing data, performs well at identifying relevant regions but often produces coarse or fragmented predictions. In contrast, SAM3 \citep{carion2025sam}, a state-of-the-art segmentation model, excels at generating high-quality masks but struggles to accurately locate target objects in large, complex images. Similar issues are observed in other models such as EarthMind \citep{shu2025earthmindleveragingcrosssensordata} and Falcon \citep{yao2025falcon}, highlighting that no single model is sufficient to handle all aspects of the grounding task effectively.

To address these challenges, we propose a unified framework that integrates the strengths of multiple models. We introduce two new pipelines—\textit{Sequential Grounding Refinement (SGR)} and \textit{Cluster-Aware Grounding Refinement (CGR)}—which combine the localization capability of RemoteSAM with the refinement power of SAM3. RemoteSAM is first used to obtain an initial estimate of object location, which is then refined using SAM3 to produce more accurate and spatially coherent results. Furthermore, we propose a majority-voting ensemble strategy that aggregates predictions from multiple complementary pipelines, including RemoteSAM, SAM3, EarthMind, Falcon, and our proposed methods.

Our main contributions are:

\begin{itemize}
\item We propose two grounding pipelines, Sequential Grounding Refinement (SGR) and Cluster-Aware Grounding Refinement (CGR), that combine localization from RemoteSAM with segmentation refinement from SAM3.
\item We introduce a clustering-based refinement strategy that improves spatial coherence of grounding predictions.
\item We propose a majority-voting ensemble across multiple grounding pipelines to improve robustness.
\item Experiments on VRS Bench and NWPU-VHR-10 demonstrate improved performance over individual models.
\end{itemize}

Our code is available at \url{https://github.com/PanavShah1/LG-SAM}.

\section{Related Work}

\paragraph{Visual Grounding.}
Visual grounding aims to localize image regions corresponding to natural language descriptions. 
Early work focused primarily on natural images, while recent approaches have begun extending 
grounding techniques to remote sensing imagery. RemoteSAM \citep{yao2025remotesam} is a recent 
model designed specifically for grounding tasks in remote sensing data, leveraging language-conditioned segmentation to identify target regions.

\paragraph{Segmentation Foundation Models.}
Large-scale segmentation models such as SAM and its variants have demonstrated strong 
generalization across diverse visual domains. SAM3 \citep{carion2025sam} extends these 
capabilities by enabling high-quality mask generation from minimal prompts and has shown 
promising performance across a variety of segmentation tasks.

\paragraph{Vision–Language Models for Remote Sensing.}
Several recent models aim to bridge vision and language understanding for remote sensing imagery. 
EarthMind \citep{shu2025earthmindleveragingcrosssensordata} is a vision–language model designed for remote sensing that leverages cross-sensor data to improve multimodal understanding.
Falcon \citep{yao2025falcon} is a vision–language model designed for remote sensing imagery that supports multiple multimodal tasks, including visual grounding.

In contrast to prior work that focuses on improving individual models, our approach combines 
multiple complementary grounding and segmentation pipelines and introduces an ensemble-based 
voting strategy to improve robustness and localization accuracy.

\section{Sequential Grounding Refinement (SGR)}

\begin{figure*}
    \centering
    \includegraphics[width=0.8\linewidth]{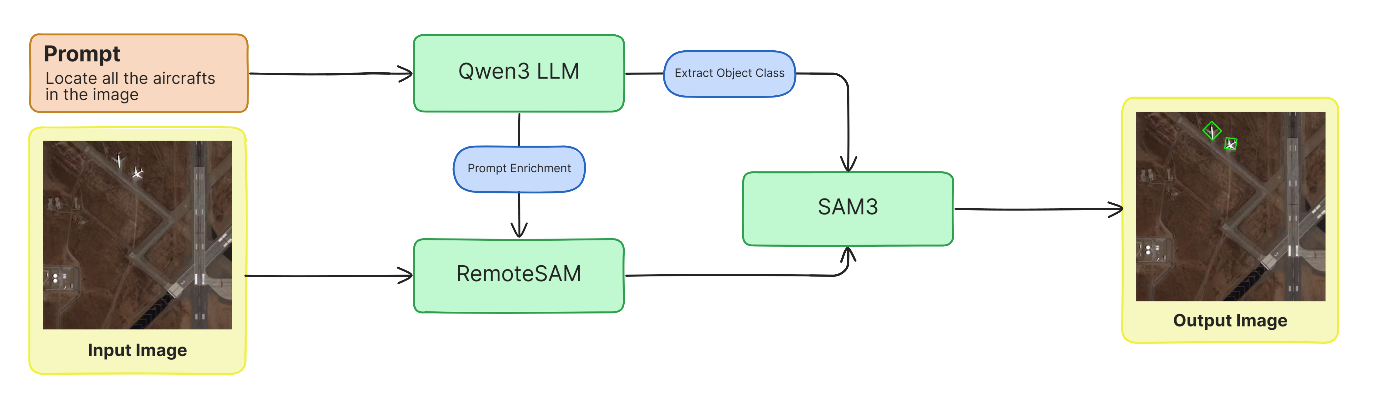}
    \caption{Architecture of Sequential Grounding Refinement (SGR)}
    \label{fig:archv1}
\end{figure*}

\label{sec:v1}

Our first proposed pipeline is designed as a simple and effective combination of RemoteSAM \citep{yao2025remotesam} and SAM3 \citep{carion2025sam}.

\subsection{Prompting}
\label{prompt_v1}
To ensure that both models receive suitable inputs, we employ Qwen3-8b \citep{yang2025qwen3technicalreport}, a large-language model, to reformulate the original user query into task-specific prompts. The original text description is provided to Qwen3-8b, which generates two separate prompts tailored to the strengths of the two models.

The first prompt retains spatial and directional information from the original query (e.g., “the yellow car on the left side of the road”). This prompt is passed to RemoteSAM, which benefits from such contextual cues when performing initial grounding. The second prompt removes all spatial and directional references and focuses solely on the target object category (e.g., “yellow car”). This simplified prompt is used as input to SAM3, along with a cropped region of the image that primarily contains the object of interest.

By generating specialized prompts for each model, this strategy allows RemoteSAM to leverage spatial context for coarse localization, while enabling SAM3 to perform precise segmentation without being confused by high-level directional language which degrades the quality of its outputs.

\subsection{RemoteSAM}
The original image and the text prompt generated by Qwen3-8b are provided as inputs to RemoteSAM. Based on the prompt, RemoteSAM may produce multiple oriented bounding box predictions for a single query, corresponding to different candidate regions in the image. Each predicted box is cropped from the original image and independently forwarded to SAM3 for refinement. 

\subsection{SAM3}
SAM3 processes the cropped regions one at a time, where each crop primarily contains the target object identified by RemoteSAM. Restricting the input to localized image crops reduces background clutter and limits the search space for the segmentation model, enabling SAM3 to generate more accurate segmentations. Given these focused inputs, SAM3 produces high-quality segmentation masks with precise object boundaries. The resulting masks are converted into oriented bounding boxes by thresholding the confidence map, extracting connected components, and fitting minimum-area rectangles using OpenCV’s \texttt{minAreaRect}. These boxes serve as the final grounding outputs of the pipeline.

Figure~\ref{fig:archv1} illustrates the architecture of this pipeline.

\section{Cluster-Aware Grounding Refinement (CGR)}

\begin{figure*}
    \centering
    \includegraphics[width=1\linewidth]{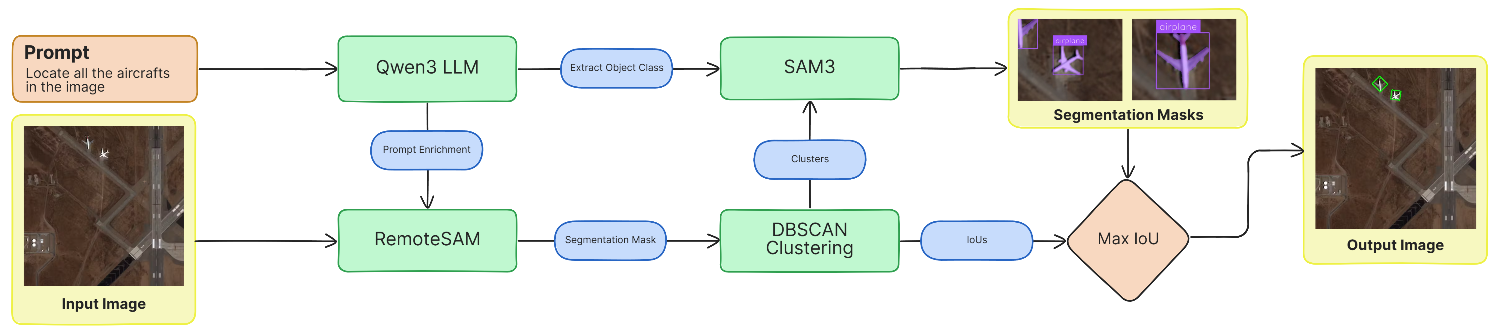}
    \caption{Architecture of Cluster-Aware Grounding Refinement (CGR)}
    \label{fig:archv2}
\end{figure*}

\label{sec:v2}
The SGR pipeline has several limitations that motivate the design of CGR. In many cases, the grounding outputs from RemoteSAM \citep{yao2025remotesam} are fragmented and contain unnecessary extra boxes or overlapping predictions for the same object. These inconsistent outputs lead to inaccurate results, as SAM3 \citep{carion2025sam} is forced to process each individual box separately, often producing redundant or incorrect segmentations. 

To address these issues, we propose a refined pipeline that reduces fragmentation and generates more reliable object candidates before passing them to SAM3. Compared to the first pipeline, this version replaces raw bounding box predictions with clustered regions derived from the RemoteSAM logit mask.

\subsection{Prompting}
We follow the same prompting procedure as used in Sequential Grounding Refinement (SGR) (Section~\ref{prompt_v1}).

\subsection{RemoteSAM and Logit Mask}

In this stage, the original image and text prompt are provided as inputs to RemoteSAM, following the same procedure as in the Sequential Grounding Refinement (SGR) pipeline. RemoteSAM typically produces multiple bounding box predictions; however, these outputs are often fragmented and may not correspond to meaningful or spatially coherent object regions.

To address this limitation, we utilize the segmentation logit mask produced by RemoteSAM instead of relying directly on its bounding box outputs. The logit mask provides dense pixel-level information that better captures the spatial structure of candidate regions. We apply DBSCAN clustering \citep{ester1996dbscan} to this mask to group related pixels into coherent object segments. This process effectively filters out noisy and fragmented predictions, retaining only the most relevant and spatially consistent regions.

DBSCAN is particularly suitable for this task because it groups spatially dense regions while automatically discarding isolated noise points. The clustering parameters are chosen to balance the separation of nearby objects with the removal of fragmented predictions.

By operating on the logit mask rather than raw bounding boxes, this approach produces more reliable object candidates that can be further refined in subsequent stages of the pipeline.
\subsection{SAM3 and Cluster Matching}
For each DBSCAN cluster, the corresponding region is cropped from the image with additional padding to ensure that the full object is included, as high-probability points may not cover the entire instance. Each cropped region is then processed by SAM3, which generates segmentation masks for all detected instances of the target object within that crop.

Since a cropped region may contain multiple objects, including instances belonging to other clusters, an additional filtering step is required. To identify the correct segmentation, we compute the Intersection-over-Union (IoU) between each SAM3-generated mask and the original cluster points. The segmentation with the highest IoU score is selected as the final output for that cluster, as it best corresponds to the intended object. This process ensures that only the most relevant and spatially consistent segmentation is retained, leading to more accurate grounding results.

Figure~\ref{fig:archv2} illustrates the architecture of this pipeline.

\section{Majority Voting}

To further improve robustness, we adopt an ensemble strategy based on majority voting across six different pipelines: RemoteSAM \citep{yao2025remotesam}, SAM3 \citep{carion2025sam}, EarthMind \citep{shu2025earthmindleveragingcrosssensordata}, Falcon \citep{yao2025falcon}, Sequential Grounding Refinement (Section~\ref{sec:v1}), and Cluster-Aware Grounding Refinement (Section~\ref{sec:v2}). Each pipeline produces a set of predicted bounding boxes for a given image–query pair. Because the pipelines rely on different architectures and training paradigms, their predictions often vary in both localization accuracy and the number of detected objects.

Rather than directly merging all predictions, we select the most reliable pipeline output by measuring its agreement with the other pipelines. Predictions that are consistent across multiple independent models are more likely to correspond to correct object localizations.

For each pipeline $i$, we compute a score that measures the similarity between its predictions and those produced by all other pipelines. This score is defined in Equation~\ref{majority}. The similarity between two pipelines $i$ and $j$ is computed using the Intersection-over-Union (IoU) between their predicted bounding boxes. When multiple boxes are present, we compute the average IoU between the best-matching pairs of boxes across the two prediction sets. This provides a measure of how spatially consistent the predictions are across models.

In addition to spatial agreement, we also account for differences in the number of predicted boxes. Some models may produce significantly more or fewer detections than others, which can indicate noisy or incomplete predictions. To mitigate this issue, we introduce an exponential penalty based on the absolute difference between the number of predicted boxes from two pipelines. This term downweights comparisons between pipelines that produce highly inconsistent detection counts.

\begin{equation}
\begin{aligned}
\text{score}[i] &= \sum_{j \neq i} \exp\!\left(-\lambda|\text{boxes}(i) - \text{boxes}(j)|\right)\cdot \text{IoU}(i,j), \\
i^{*} &= \arg\max_i \ \text{score}[i]
\end{aligned}
\label{majority}
\end{equation}

The final score for each pipeline therefore reflects both spatial alignment and consistency in prediction counts. Pipelines that generate bounding boxes similar to those of other models while maintaining a reasonable number of detections receive higher scores. After computing the score for each pipeline, the pipeline with the highest score is selected, and its predicted bounding boxes are used as the final grounding result.

This ensemble-based approach leverages consensus among multiple complementary grounding models while avoiding the complexity of directly merging heterogeneous predictions. As a result, it improves robustness to individual model failures and produces more stable grounding outputs across diverse remote sensing scenes.

\begin{table*}[h!]
\centering
\begin{tabular}{p{10.5cm} c c c c}
\hline
\textbf{Model} & \textbf{mIoU} & \textbf{Acc@0.5} & \textbf{Acc@0.7} & \textbf{Avg Cnt Diff}  \\
\hline
RemoteSAM & 0.6283 & 0.7812 & 0.3931 & $-0.0803$  \\
SAM3 & 0.1635 & 0.2051 & 0.0887 & 0.3060 \\
EarthMind & 0.5961 & 0.7414 & 0.3732 & $-0.0879$  \\
Falcon & 0.5802 & 0.7221 & 0.4709 & $-1.3629$  \\
Sequential Grounding Refinement (SGR) & 0.5671 & 0.6827 & 0.3303 & $-0.1118$  \\
Cluster-Aware Grounding Refinement (CGR) & 0.6315 & 0.7807 & 0.3909 & $-0.2454$  \\
RemoteSAM + SAM3 + Sequential Grounding Refinement (SGR) & 0.6325 & 0.7871 & 0.3936 & $-0.2277$  \\
RemoteSAM + SAM3 + Cluster-Aware Grounding Refinement (CGR) & 0.6318 & 0.7870 & 0.3948 & $-0.2715$ \\
RemoteSAM + SAM3 + EarthMind & 0.5431 & 0.6713 & 0.3519 & $-0.2091$   \\
RemoteSAM + Sequential Grounding Refinement (SGR) + EarthMind & 0.6359 & 0.7958 & 0.4047 & $-0.1683$  \\
RemoteSAM + Sequential Grounding Refinement (SGR) + Falcon & 0.6430 & 0.7978 & 0.4217 & $-0.2380$ \\
RemoteSAM + SAM3 + Falcon & 0.6352 & 0.7899 & 0.4042 & $-0.3636$\\
RemoteSAM + Cluster-Aware Grounding Refinement (CGR) + EarthMind & 0.6297 & 0.7823 & 0.4080 & $-0.0958$\\
RemoteSAM + SAM3 + Sequential Grounding Refinement (SGR) + EarthMind & 0.6377 & 0.7988 & 0.4049 & $-0.0958$   \\
RemoteSAM + SAM3 + Cluster-Aware Grounding Refinement (CGR) + EarthMind & 0.6282 & 0.7826 & 0.3988 & $-0.2522$  \\
\makecell[l]{RemoteSAM + SAM3 + Sequential Grounding Refinement (SGR)\\ + Cluster-Aware Grounding Refinement (CGR) + EarthMind + Falcon \\(Majority Voting across all models)} & 0.6494 & 0.7928 & 0.4121 & 0.1408 \\

\hline
\end{tabular}
\caption{Evaluation of different model combinations (VRS Bench dataset \citep{li2024vrsbench}). Here, `+' denotes the majority voting of the models.}
\label{tab:model_results_vrs}
\end{table*}

\begin{table*}[h]
\centering
\begin{tabular}{p{10.5cm} c c c c}
\hline
\textbf{Model} & \textbf{mIoU} & \textbf{Acc@0.5} & \textbf{Acc@0.7} & \textbf{Avg Cnt Diff} \\
\hline
RemoteSAM & 0.5662 & 0.6624 & 0.2484 & $-1.4825$\\
SAM3 & 0.5009 & 0.6352 & 0.1683 & $-1.2361$ \\
EarthMind & 0.5256 & 0.6088 & 0.1844 & $-1.1916$ \\
Falcon & 0.5415 & 0.6271 & 0.3051 & $1.0584$ \\
Sequential Grounding Refinement (SGR) & 0.5098 & 0.6057 & 0.1972 & 3.2265 \\
Cluster-Aware Grounding Refinement (CGR) & 0.5719 & 0.6953 & 0.2686 & $-1.4253$ \\
RemoteSAM + SAM3 + Sequential Grounding Refinement (SGR) & 0.5755 & 0.6851 & 0.2844 & $-0.5895$ \\
RemoteSAM + SAM3 + Cluster-Aware Grounding Refinement (CGR) & 0.6321 & 0.7219 & 0.4051 & $-0.2522$ \\
RemoteSAM + SAM3 + EarthMind & 0.5908 & 0.7426 & 0.2416 & $-0.2882$ \\
RemoteSAM + Sequential Grounding Refinement (SGR) + EarthMind & 0.5711 & 0.6859 & 0.2634 & $-0.7101$ \\
RemoteSAM + Sequential Grounding Refinement (SGR) + Falcon & 0.5787 & 0.6924 & 0.2733 & $-0.6021$ \\
RemoteSAM + SAM3 + Falcon & 0.5909 & 0.7335 & 0.2485 & $-0.4109$ \\
RemoteSAM + Cluster-Aware Grounding Refinement (CGR) + EarthMind & 0.5933 & 0.7277 & 0.3022 & 0.2332 \\
RemoteSAM + SAM3 + Sequential Grounding Refinement (SGR) + EarthMind & 0.5928 & 0.7471 & 0.2468 & $-1.1917$ \\
RemoteSAM + SAM3 + Cluster-Aware Grounding Refinement (CGR) + EarthMind & 0.5924 & 0.7467 & 0.2572 & $0.6744$ \\
\makecell[l]{RemoteSAM + SAM3 + Sequential Grounding Refinement (SGR)\\ + Cluster-Aware Grounding Refinement (CGR) + EarthMind + Falcon \\(Majority Voting across all models)} & 0.6031 & 0.7293 & 0.3329 & 0.9842 \\ 
\hline
\end{tabular}
\caption{Evaluation of different model combinations (NWPU-VHR-10 dataset \citep{Cheng_2016}). Here, `+' denotes the majority voting of the models}
\label{tab:model_results_vhr}
\end{table*}

\section{Evaluation Setup}

\subsection{Datasets}
\label{sec:data}

\begin{figure*}[t]
\centering
\begin{tabular}{cc}
\includegraphics[width=0.35\textwidth]{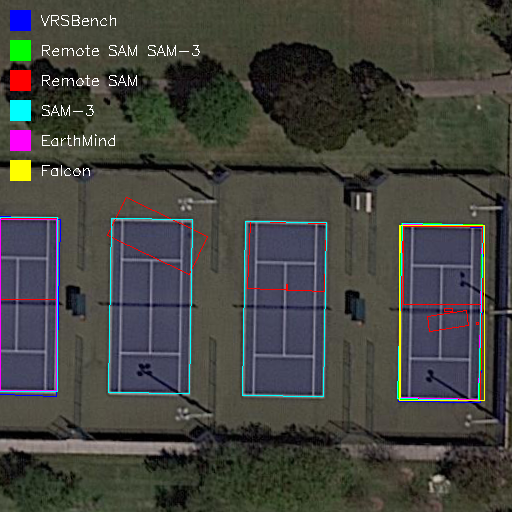} &
\includegraphics[width=0.35\textwidth]{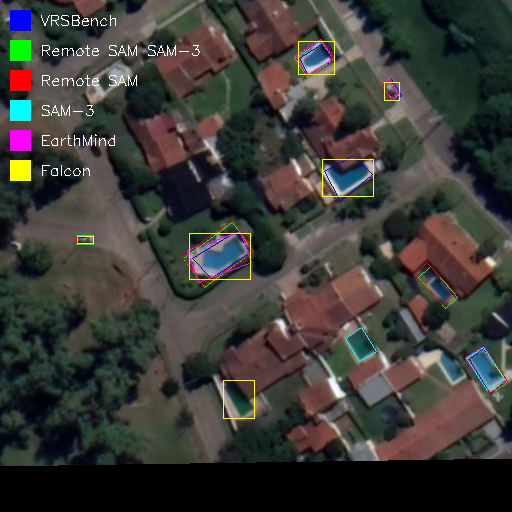} \\
\includegraphics[width=0.35\textwidth]{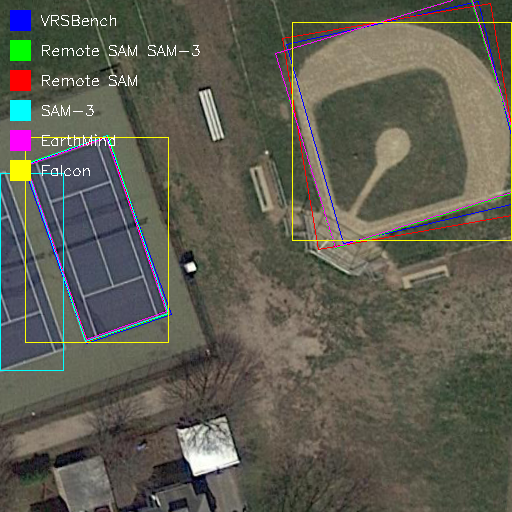} &
\includegraphics[width=0.35\textwidth]{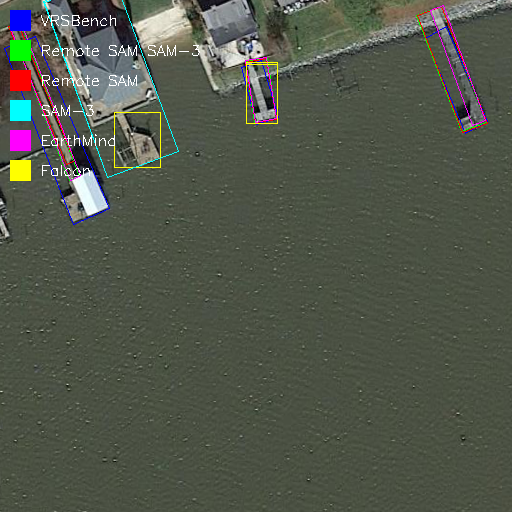} \\
\end{tabular}
\caption{Grounding results from multiple models on images from VRS Bench \citep{li2024vrsbench}. Colors denote: ground truth (blue), Cluster-Aware Grounding Refinement (CGR) (green), RemoteSAM (red), SAM3 (light blue), EarthMind (pink), and Falcon (yellow).}

\end{figure*}

\begin{itemize}
    \item VRS Bench \citep{li2024vrsbench}: VRS Bench is a large-scale, multi-task vision–language benchmark for remote sensing which can support captioning, VQA, and visual grounding tasks. It contains 29614 images, 29614 human-verified captions, 52472 referring-expression annotations, and 123221 VQA pairs. Each image has resolution $512 \times 512$.  It is built through a semi-automatic pipeline combining object-level metadata extraction, LLM-based annotation generation, and human verification, resulting in high-quality, fine-grained annotations. 
    \item NWPU-VHR-10 \citep{Cheng_2016}: NWPU-VHR-10 is a very-high-resolution remote sensing object-detection dataset designed for evaluating detection algorithms in complex urban and industrial scenes. It contains 800 images, including 650 positive images with 3651 annotated objects and 150 negative images. The dataset covers 10 categories: airplane, ship, storage tank, baseball diamond, tennis court, basketball court, ground track field, harbor, bridge, and vehicle. Images have spatial resolutions ranging from 0.2 m to 1.0 m, with varying image sizes. All instances are annotated using axis-aligned bounding boxes, providing high-quality labels for benchmarking small-object detection and multi-scale recognition in VHR remote sensing imagery.
\end{itemize}

\subsection{Metrics}
\label{eval_metrics}

We evaluate grounding performance using Intersection-over-Union (IoU) between the predicted region $P$ and ground-truth region $G$:

\begin{equation}
IoU = \frac{|P \cap G|}{|P \cup G|}.
\end{equation}

We report the mean Intersection-over-Union (mIoU) across all $N$ evaluation samples:

\begin{equation}
mIoU = \frac{1}{N} \sum_{i=1}^{N} IoU_i,
\end{equation}

where $IoU_i$ denotes the IoU for the $i$-th image–query pair. In addition, we report Acc@0.5 and Acc@0.7, which measure the percentage of predictions whose IoU with the ground-truth bounding box exceeds 0.5 and 0.7 respectively.

\subsection{Hyperparameters Used}

All models were evaluated using pretrained checkpoints without additional training. 
In the CGR pipeline, the RemoteSAM logit mask was thresholded by exponentially reducing the confidence threshold starting from $0.99$ and decaying by $10\%$ on every iteration until at least 50 points were included in the mask. These points were then clustered 
using DBSCAN with parameters $\epsilon = 20.0$ and min\_samples $ = 5$. Each cluster was cropped with additional padding of $10\%$ of the cropped region and processed by SAM3. The final mask for each cluster was selected based on the maximum IoU with the cluster pixels. For the ensemble pipeline, the penalty term $\exp{(-\lambda|n_i - n_j|)}$ was applied to account for differences in detection counts, with $\lambda = 0.5$.

\section{Results}
\label{sec:results}

Tables~\ref{tab:model_results_vrs} and \ref{tab:model_results_vhr} present quantitative results on the VRS Bench \citep{li2024vrsbench} and NWPU-VHR-10 datasets \citep{Cheng_2016} respectively. Figure~\ref{fig:grounding_pipelines} illustrates qualitative examples from different grounding pipelines. 

SAM3 alone often struggles to accurately localize target objects in cluttered scenes, resulting in a high average count difference. RemoteSAM generally produces reasonable predictions, but its outputs can be fragmented or coarse in challenging cases. EarthMind and Falcon perform well in specific scenarios but remain inconsistent across diverse queries and object categories.

The Sequential Grounding Refinement (SGR) pipeline performs worse than RemoteSAM alone, indicating that a straightforward combination of the two models does not reliably improve performance. In contrast, the improved Cluster-Aware Grounding Refinement (CGR) pipeline surpasses RemoteSAM, demonstrating the effectiveness of the proposed refinement strategy. Furthermore, the majority-voting ensemble achieves the best performance, improving mIoU over RemoteSAM by 6.5\% on the NWPU-VHR-10 dataset and by 3.4\% on the VRS Bench dataset.

\section{Discussion}

The results demonstrate that different grounding models exhibit complementary strengths and weaknesses. While RemoteSAM provides reliable localization, its outputs can be fragmented in challenging scenes. SAM3 produces high-quality segmentations but often struggles to identify the correct object location in large remote sensing images. The proposed Cluster-Aware Grounding Refinement (CGR) pipeline addresses these limitations by combining coarse localization with segmentation refinement, resulting in improved grounding performance.

The ensemble strategy further improves robustness by leveraging agreement across multiple pipelines. By selecting predictions that are most consistent across models, the majority-voting framework reduces the impact of individual model errors and improves overall localization accuracy.

However, the majority-voting framework has certain limitations. In particular, the ensemble is currently biased toward RemoteSAM, as three out of the six pipelines incorporate RemoteSAM as a core component. This may lead to an overrepresentation of its predictions in the final output, potentially reducing the diversity of the ensemble. In future work, we plan to incorporate additional independent grounding models to create a more balanced set of pipelines. We also aim to explore weighted voting schemes that assign different importance to each model based on reliability, which could further reduce bias and improve robustness.

The majority-voting framework also substantially increases the computational requirements, as six independent pipelines are executed to generate the final output. However, this increase in compute does not affect the overall runtime of the framework, since the pipelines are mutually independent and can be executed in parallel without interdependencies.

\section{Conclusion}

This work proposes two grounding pipelines that combine the localization capabilities of RemoteSAM with the segmentation strength of SAM3. The improved Cluster-Aware Grounding Refinement (CGR) pipeline produces more coherent object candidates and achieves better grounding accuracy. In addition, a majority-voting ensemble across multiple grounding pipelines further improves robustness. Experiments on the VRS Bench and NWPU-VHR-10 datasets demonstrate that integrating complementary models leads to more reliable visual grounding in remote sensing imagery.

\newpage
{
    \small
    \bibliographystyle{IEEEtran}
    \bibliography{main}
}


\end{document}